\documentclass{article}

\usepackage{microtype}
\usepackage{multirow}
\usepackage{graphicx}
\usepackage{subcaption}
\usepackage{booktabs} 
\usepackage{hyperref}
\usepackage{enumitem}

\usepackage{amsmath}
\usepackage{amssymb}
\usepackage{mathtools}
\usepackage{amsthm}

\usepackage{placeins}

\usepackage{algorithm}
\usepackage{algorithmic}

\usepackage{natbib}

\usepackage[capitalize,noabbrev]{cleveref}

\usepackage[preprint]{icml2026}

\theoremstyle{plain}

\theoremstyle{definition}

\theoremstyle{remark}

\usepackage[textsize=tiny]{todonotes}

\icmltitlerunning{Co2PO: Coordinated Constrained Policy Optimization for Multi-Agent RL}

\begin{document}

\twocolumn[
\icmltitle{Co2PO: Coordinated Constrained Policy Optimization for Multi-Agent RL}

\icmlsetsymbol{equal}{$\dagger$}

\begin{icmlauthorlist}
  \icmlauthor{Shrenik Patel}{equal,rutgers}
  \icmlauthor{Christine Truong}{equal,cmu}
\end{icmlauthorlist}

\icmlaffiliation{rutgers}{Rutgers University, New Brunswick, NJ}
\icmlaffiliation{cmu}{Carnegie Mellon University, Pittsburgh, PA}

\icmlcorrespondingauthor{Shrenik Patel}{shrenik.patel@rutgers.edu}
\icmlcorrespondingauthor{Christine Truong}{cmtruong@andrew.cmu.edu}

\icmlkeywords{Constrained Reinforcement Learning, Multi-Agent Reinforcement Learning, Communication, Safety}

\vskip 0.3in
]

\printAffiliationsAndNotice{\icmlEqualContribution}  

\begin{abstract}
Constrained multi-agent reinforcement learning (MARL) faces a fundamental tension between exploration and safety-constrained optimization. Existing leading approaches, such as Lagrangian methods, typically rely on global penalties or centralized critics that react to violations after they occur, often suppressing exploration and leading to over-conservatism. We propose Co2PO, a novel MARL communication-augmented framework that enables coordination-driven safety through selective, risk-aware communication. Co2PO introduces a shared blackboard architecture for broadcasting positional intent and yield signals, governed by a learned hazard predictor that proactively forecasts potential violations over an extended temporal horizon. By integrating these forecasts into a constrained optimization objective, Co2PO allows agents to anticipate and navigate collective hazards without the performance trade-offs inherent in traditional reactive constraints. We evaluate Co2PO across a suite of complex multi-agent safety benchmarks, where it achieves higher returns compared to leading constrained baselines while converging to cost-compliant policies at deployment. Ablation studies further validate the necessity of risk-triggered communication, adaptive gating, and shared memory components.
\end{abstract}

\section{Introduction}
\label{sec:intro}

Constrained multi-agent reinforcement learning (MARL) aims to maximize collective return while satisfying safety constraints, but in cooperative settings this tension is especially sharp: aggressive exploration can incur costly violations, while overly conservative updates can prematurely suppress learning and prevent agents from discovering high-performing joint behavior \cite{brunke2022safelearning}. The problem is amplified by partial observability and interaction-driven dynamics. Even when agents share the same reward objective, safety failures often emerges from coordination breakdowns in timing, intent, or mutual awareness, rather than from conflicting incentives \cite{omidshafiei2015decentralizedcontrolpartiallyobservable, calzolari2026safeheterogeneousmultiagentrl}. As a result, purely reactive constraint handling can be slow to correct the underlying cause of violations, while heavy-handed penalties can make learning brittle \cite{brunke2022safelearning}.

We introduce Co2PO, a coordinated constrained policy optimization framework that targets these coordination-driven safety failures through \emph{selective, risk-aware coordination}. Rather than treating communication as a persistent channel, Co2PO uses it as a sparse control mechanism: agents write to a shared blackboard only when they predict elevated near-term hazard risk. Each write contains compact safety-relevant signals---a positional state summary, an intent vector, and a yield flag---that other agents can retrieve to condition their actions. This design preserves decentralized action selection while providing coordination signals precisely when interaction risk is most acute.

Across cooperative multi-agent safety benchmarks, Co2PO improves feasible performance and reaches cost-compliant policies at evaluation under the same training budget. We observe transient violations early in training, when hazard prediction and coordination behavior are still being learned, but final policies satisfy the cost constraint during evaluation. Our contributions are as follows:

\noindent\textbf{Contributions.}\par
\begin{itemize}\setlength\topsep{0pt}\setlength\itemsep{5pt}\setlength\parskip{0pt}\setlength\parsep{0pt}
  \item We introduce Co2PO, enabling selective coordination via a shared blackboard with decentralized actions.
  \item We propose hazard-triggered writes of compact signals (state, intent, yield) with adaptive write-rate control.
  \item We evaluate on cooperative multi-agent safety benchmarks with ablations isolating blackboard and gating.
\end{itemize}


\section{Problem Setup and Preliminaries}
\label{sec:setup}

We study constrained cooperative multi-agent reinforcement learning (MARL) under partial observability. Agents seek high return while satisfying an expected cost budget at evaluation, and we summarize performance with \emph{feasible return}.

\subsection{Constrained Cooperative Markov Game}
We model the environment as a cooperative constrained Markov game (CMG) with $n$ agents \cite{littman1994markovgames}:
\[
\mathcal{G} =
\langle
\mathcal{N}, \mathcal{S}, \{\mathcal{O}_i\}_{i\in\mathcal{N}},
\{\mathcal{A}_i\}_{i\in\mathcal{N}}, P, r, c, \gamma, \rho_0, d
\rangle .
\]
Agent $i$ observes $o_t^i\in\mathcal{O}_i$ and selects $a_t^i\in\mathcal{A}_i$, yielding joint action $a_t=(a_t^1,\dots,a_t^n)$ and transition $s_{t+1}\sim P(\cdot\mid s_t,a_t)$ from $s_0\sim\rho_0$.
The environment emits a shared reward $r_t=r(s_t,a_t)$ and safety cost $c_t=c(s_t,a_t)$.

Policies factorize as $\pi(a\mid o)=\prod_{i=1}^n \pi_i(a^i\mid o^i)$ with $o=(o^1,\dots,o^n)$.
The discounted return and cost are
\begin{align}
J_R(\pi) &= \mathbb{E}_{\pi}\!\left[\sum_{t=0}^{\infty}\gamma^t\, r(s_t,a_t)\right], \label{eq:JR}\\
J_C(\pi) &= \mathbb{E}_{\pi}\!\left[\sum_{t=0}^{\infty}\gamma^t\, \bar{c}(s_t,a_t)\right], \label{eq:JC}
\end{align}
where $\bar{c}$ is the team cost used for enforcement.
We seek
\begin{equation}
\max_{\pi}\; J_R(\pi)\quad \text{s.t.}\quad J_C(\pi)\le d,
\label{eq:cmdg}
\end{equation}
with cost budget $d$.

\subsection{CTDE Preliminaries}
\label{sec:ctde}
We use centralized training with decentralized execution (CTDE): critics may condition on centralized information during training, while each actor $\pi_i$ conditions only on its local observation (and any permitted message/memory) at execution \cite{lowe2020multiagentactorcriticmixedcooperativecompetitive, yu2022surprisingeffectivenessppocooperative}.

\subsection{Safety Metric: Feasible Return}
\label{sec:safety-metrics}
We report \emph{feasible return} to reflect deployment-time cost constraints:
\begin{equation}
J_F(\pi) \;=\; J_R(\pi)\cdot \mathbf{1}\!\left[J_C(\pi)\le d\right].
\label{eq:feasible}
\end{equation}

\subsection{Lagrangian Relaxation}
\label{sec:lagrangian-prelims}
We optimize a Lagrangian relaxation of \cref{eq:cmdg}, following standard constrained policy optimization formulations \cite{achiam2017constrainedpolicyoptimization}:
\begin{equation}
\max_{\pi}\; \min_{\lambda\ge 0}\;\; \mathcal{L}(\pi,\lambda)
\;=\;
J_R(\pi)\;-\;\lambda\big(J_C(\pi)-d\big),
\label{eq:lagrangian}
\end{equation}
with the dual variable $\lambda$ updated online. Actor--critic updates commonly use a hybrid advantage \cite{tessler2018rcpo, ray2019safexp, zhang2020orderconstrainedoptimizationpolicy}
\begin{equation}
A^{\text{hyb}}_t \;=\; A^{R}_t \;-\; \lambda\, A^{C}_t.
\label{eq:hybrid-adv}
\end{equation}

\subsection{Risk Events and Hazard Labels}
\label{sec:hazard-labels}
We construct a proactive hazard label from dense per-step costs. Let $c_t^i$ be the instantaneous cost for agent $i$.
Define a hazard event
\begin{equation}
z_t^i \;=\; \mathbf{1}\!\left[c_t^i > \delta\right],
\label{eq:inst-hazard}
\end{equation}
and a lookahead label with horizon $H$,
\begin{equation}
h_t^i
\;=\;
\mathbf{1}\!\left[\max_{\tau \in \{t,\dots,t+H\}} z_\tau^i = 1\right],
\label{eq:lookahead-hazard}
\end{equation}
computed within episode boundaries. We use $h_t^i$ to supervise the hazard predictor used for selective coordination in \cref{sec:method}.

\section{Methodology}
\label{sec:method}

We present Co2PO, a constrained MARL method that augments decentralized policies with \emph{selective} shared-memory coordination.
Each agent predicts near-term hazard risk from its local observation and writes a compact message to a shared blackboard only when predicted risk exceeds a threshold.
Agents then read a fixed-size context from the blackboard and condition their actions on the retrieved information.
Training uses a constrained actor--critic objective with reward and cost critics, a dual variable for constraint enforcement, supervised hazard forecasting, and a penalty that discourages excessive writing.

\subsection{Per-Timestep Interaction Protocol}
\label{sec:protocol}
We consider $E$ parallel environment instances and $n$ agents. At timestep $t$ in environment $e$:
\emph{(1) write-info} each agent computes hazard probability and message contents from its observation;
\emph{(2) write} agents gate writes and update the blackboard;
\emph{(3) read} each agent retrieves up to $k$ other agents' messages to form a fixed-length context;
\emph{(4) act} each agent selects an action conditioned on its observation and retrieved context.
The blackboard is maintained per environment instance and cleared on episode termination.

\subsection{Blackboard: State, Write, and Read}
\label{sec:blackboard}

\paragraph{Stored fields.}
For each environment $e$ and agent $i$, the blackboard stores a single current entry
\begin{equation}
B_t^{e,i} \;=\; \big(x_t^{e,i},\, u_t^{e,i},\, y_t^{e,i},\, p_t^{e,i},\, w_t^{e,i}\big),
\label{eq:blackboard-entry}
\end{equation}
where $x_t^{e,i}\in\mathbb{R}^{d}$ is a compact state summary,
$u_t^{e,i}\in\mathbb{R}^{d}$ is a learned intent vector,
$y_t^{e,i}\in[0,1]$ is a learned yield flag,
$p_t^{e,i}\in[0,1]$ is the hazard probability,
and $w_t^{e,i}\in\{0,1\}$ is the write indicator.

\paragraph{Write information from local observation.}
Each agent computes message fields and a hazard score from its observation:
\begin{equation}
(\ell_t^i,\; u_t^i,\; y_t^i) \;=\; F_\theta(o_t^i),
\qquad
p_t^i \;=\; \sigma(\ell_t^i),
\label{eq:write-info}
\end{equation}
where $\ell_t^i$ is a hazard logit and $\sigma(\cdot)$ is the sigmoid.

\paragraph{Event-triggered writes.}
Writes are gated by a threshold:
\begin{equation}
w_t^{e,i} \;=\; \mathbf{1}\!\left[p_t^{e,i} > \tau_t\right],
\label{eq:gate}
\end{equation}
and only entries with $w_t^{e,i}=1$ are valid for reads.


\paragraph{Similarity-based read (top-$k$) with fixed-length context.} At each timestep, agent $i$ forms a query vector from its own state summary, $q_t^{e,i}=x_t^{e,i}$.
It then considers the set of \emph{active} entries written by other agents in the same environment instance,
$\{\,B_t^{e,j}\,:\, j\neq i,\; w_t^{e,j}=1\,\}$, and scores each candidate $j$ by cosine similarity \cite{das2020tarmactargetedmultiagentcommunication}:
\[
s_t^{e}(i,j)=\left\langle
\frac{x_t^{e,j}}{\lVert x_t^{e,j}\rVert_2+\varepsilon},\;
\frac{q_t^{e,i}}{\lVert q_t^{e,i}\rVert_2+\varepsilon}
\right\rangle,
\]
where $\varepsilon>0$ is a small constant for numerical stability.
Agent $i$ retrieves the $k$ highest-scoring active entries and, for each retrieved sender $j$, constructs
\[
\psi(B_t^{e,j})=\big[x_t^{e,j};\,u_t^{e,j};\,y_t^{e,j};\,p_t^{e,j}\big]\in\mathbb{R}^{2d+2}.
\]
The memory context is the concatenation of up to $k$ such vectors in ranked order, zero-padded if fewer than $k$ entries are active:
\begin{equation}
m_t^{e,i}=\big[\psi(B_t^{e,j_1});\dots;\psi(B_t^{e,j_k})\big]\in\mathbb{R}^{k(2d+2)}.
\label{eq:memctx}
\end{equation}

\subsection{Threshold Control (Optional)}
\label{sec:adaptive-threshold}
We allow adaptation of $\tau_t$ online to target a desired write rate $\rho^\star$. Let $\rho_t$ be the observed mean write rate and $\bar{\rho}_t$ its exponential moving average \cite{stooke2020responsive}:
\begin{align}
\bar{\rho}_t &= \beta\,\bar{\rho}_{t-1} + (1-\beta)\,\rho_t, \label{eq:ema}\\
\tau_{t+1} &=
\operatorname{clip}\!\Big(\tau_t + \eta(\bar{\rho}_t-\rho^\star),\; \tau_{\min},\; \tau_{\max}\Big),
\label{eq:thr-update}
\end{align}
where $\beta\in(0,1)$, step size $\eta>0$, and bounds $(\tau_{\min},\tau_{\max})$ prevent degenerate write thresholds.

\subsection{Policy and Critics}
\label{sec:policy-critics}

\paragraph{Memory-conditioned decentralized actor.}
Each actor conditions on its observation and the retrieved context:
\begin{equation}
a_t^i \sim \pi_{\phi_i}\!\left(\cdot \mid o_t^i,\; \mathrm{enc}(m_t^i)\right),
\label{eq:actor}
\end{equation}
where $\mathrm{enc}(\cdot)$ maps \cref{eq:memctx} to an embedding concatenated with $o_t^i$.

\paragraph{Centralized critics.}
Under CTDE, we learn reward and cost critics $V^R(\cdot)$ and $V^C(\cdot)$ to compute reward and cost advantages for policy optimization.

\subsection{Learning Objective}
\label{sec:objective}

\paragraph{Constrained policy optimization.}
We optimize the Lagrangian objective (\cref{eq:lagrangian}) using a hybrid advantage
$A_t^{\text{hyb},i} = A_t^{R,i} - \lambda A_t^{C,i}$,
with $\lambda\ge 0$ updated by projected primal--dual learning.

\paragraph{Actor loss.}
Let $r_t^i(\phi_i)$ denote the importance ratio between current and behavior policies.
We use a clipped surrogate objective augmented with hazard supervision and a write penalty \cite{schulman2017proximalpolicyoptimizationalgorithms}:
\begin{equation}
\label{eq:actor-total}
\begin{aligned}
\mathcal{L}_{\text{actor}}^i
&=
\mathcal{L}_{\text{clip}}^i\!\left(A^{\text{hyb},i}\right)
+
\alpha_{\text{write}}\,
\mathbb{E}\!\left[w_t^i\right]
\\
&\quad
+
\alpha_{\text{haz}}\,
\mathbb{E}\!\left[
\operatorname{WBCE}\!\left(\ell_t^i,\; h_t^i\right)
\right]
-
\alpha_{\text{ent}}\,
\mathcal{H}\!\left(\pi_{\phi_i}\right).
\end{aligned}
\end{equation}
Here $\operatorname{WBCE}$ is weighted binary cross-entropy, $\mathcal{H}$ is policy entropy, and $\alpha_{\text{write}},\alpha_{\text{haz}},\alpha_{\text{ent}}\ge 0$ are coefficients.



\paragraph{Hazard supervision, critics, and dual update.}
We supervise the hazard head with the lookahead label $h_t^i$ (\cref{sec:hazard-labels}). Reward and cost critics are trained by value regression to their respective returns, and the dual variable $\lambda$ is updated via a projected step driven by observed constraint violations.

\subsection{Algorithm}
\label{sec:algorithm}

\begin{algorithm}[h!]
\caption{\textsc{Co2PO}: one rollout step at timestep $t$ (for all envs $e$ and agents $i$).}
\label{alg:co2po}
\begin{algorithmic}[1]
\STATE \textbf{Write-info:} compute $(\ell_t^{e,i},p_t^{e,i},u_t^{e,i},y_t^{e,i})$ from $o_t^{e,i}$ for all $(e,i)$
\STATE \textbf{Gate \& write:} set $w_t^{e,i} \leftarrow \mathbf{1}[p_t^{e,i}>\tau_t]$ and update blackboard entries $B_t^{e,i}$
\STATE \textbf{Adapt:} optionally update $\tau_t$ using \cref{eq:ema,eq:thr-update}
\STATE \textbf{Read:} for each $(e,i)$, form $m_t^{e,i}$ by top-$k$ retrieval from active entries (\cref{eq:memctx})
\STATE \textbf{Act:} sample $a_t^{e,i}\sim \pi_{\phi_i}(\cdot\mid o_t^{e,i}, \mathrm{enc}(m_t^{e,i}))$ and step environments
\STATE \textbf{Store:} add transitions (including $m_t^{e,i}$ and $w_t^{e,i}$) to buffer; clear blackboard for terminated envs
\end{algorithmic}
\end{algorithm}

\section{Experimental Setup}
\label{sec:experiments}

\subsection{Benchmarks}
We evaluate on cooperative multi-agent safety benchmarks from the \textsc{SafePO} suite, which builds on \textsc{Safety-Gymnasium} and provides a consistent multi-agent interface \citep{ji2024safetygymnasiumunifiedsafereinforcement}.
Each task emits a per-step reward and safety cost; our goal is to maximize return while meeting an episodic cost budget.

\paragraph{Velocity family.}
In multi-agent velocity tasks, agents must coordinate to achieve velocity-tracking objectives while avoiding unsafe behavior under task-specific safety costs \citep{6386109}.
These tasks are coordination-stressing because agents share a space: one agent’s behavior can perturb others’ motion, making it harder to maintain target speeds without constraint violations.

\paragraph{MultiGoal family.}
In multi-agent MultiGoal tasks, agents navigate shared spaces with safety costs arising from unsafe states or contacts (e.g., hazards) \citep{ji2024safetygymnasiumunifiedsafereinforcement}.
They stress partial observability, interaction-driven hazards, and implicit negotiation to remain cost-compliant.

\subsection{Baselines}
We compare against constrained MARL baselines implemented in \textsc{SafePO}: \emph{MAPPO} \cite{yu2022surprisingeffectivenessppocooperative}, \emph{HAPPO} \cite{kuba2022happo}, \emph{MACPO} \cite{gu2023macpo}, and \emph{MAPPO-Lagrangian} \cite{gu2023macpo}.
All methods are run under a matched training budget and evaluation protocol. When available, we use \textsc{SafePO}'s standard hyperparameter configurations for each baseline and environment to avoid unfair retuning. All hyperparameters for all methods (including method-specific settings) are reported in Appendix \ref{app:hparams} for reproducibility.

\subsection{Metrics}
Let an episode have horizon $T$ with $n$ agents.
Following the \textsc{SafePO} convention, we aggregate reward and cost across agents by averaging per-step signals over agents and summing over time:
\begin{align}
R \;=\; \sum_{t=0}^{T-1}\frac{1}{n}\sum_{i=1}^n r_t^i,
\qquad
C \;=\; \sum_{t=0}^{T-1}\frac{1}{n}\sum_{i=1}^n c_t^i.
\label{eq:episodic-return-cost}
\end{align}
We evaluate policies at checkpoints $t\in\mathcal{T}$ during training; let $\widehat{R}_t,\widehat{C}_t$ denote mean episodic return/cost over evaluation episodes at checkpoint $t$.
We report:
\begin{itemize}\setlength\itemsep{2pt}\setlength\parskip{0pt}\setlength\parsep{0pt}\setlength\topsep{2pt}
  \item \textbf{Final return/cost} ($R_{\text{final}},C_{\text{final}}$): $\widehat{R}_{t_{\text{last}}}$ and $\widehat{C}_{t_{\text{last}}}$ at the last checkpoint $t_{\text{last}}$.
  \item \textbf{Peak cost} ($C_{\text{peak}}$): $\max_{t\in\mathcal{T}}\widehat{C}_t$, capturing the worst evaluated cost over training.
  \item \textbf{Violation rate}: fraction of evaluation episodes whose episodic cost exceeds the budget $d$ (i.e., $\Pr[C>d]$).
  \item \textbf{Feasible return} ($R_{\mathrm{feas}}$): best evaluation return among checkpoints that satisfy the cost budget,
  \begin{equation}
    R_{\mathrm{feas}} \;=\; \max_{t \in \mathcal{T}}\Big\{\widehat{R}_t \;:\; \widehat{C}_t \le d\Big\}.
    \label{eq:feasible-return}
  \end{equation}
  \item \textbf{Time-to-feasible}: the earliest training step corresponding to a checkpoint $t$ with $\widehat{C}_t \le d$; if no checkpoint is feasible, we report ``--''.
\end{itemize}

\subsection{Training and Evaluation Protocol}
We follow the standard \textsc{SafePO} on-policy training loop: collect rollouts with $E$ parallel environment instances, perform multiple epochs of minibatch updates per iteration, and periodically evaluate the current policy in separate evaluation rollouts \citep{ji2024safetygymnasiumunifiedsafereinforcement}.
Evaluation uses deterministic action selection and reports means (± standard deviation) over three independent random seeds.

\begin{figure*}[h!]
  \centering
  \includegraphics[width=0.99\textwidth]{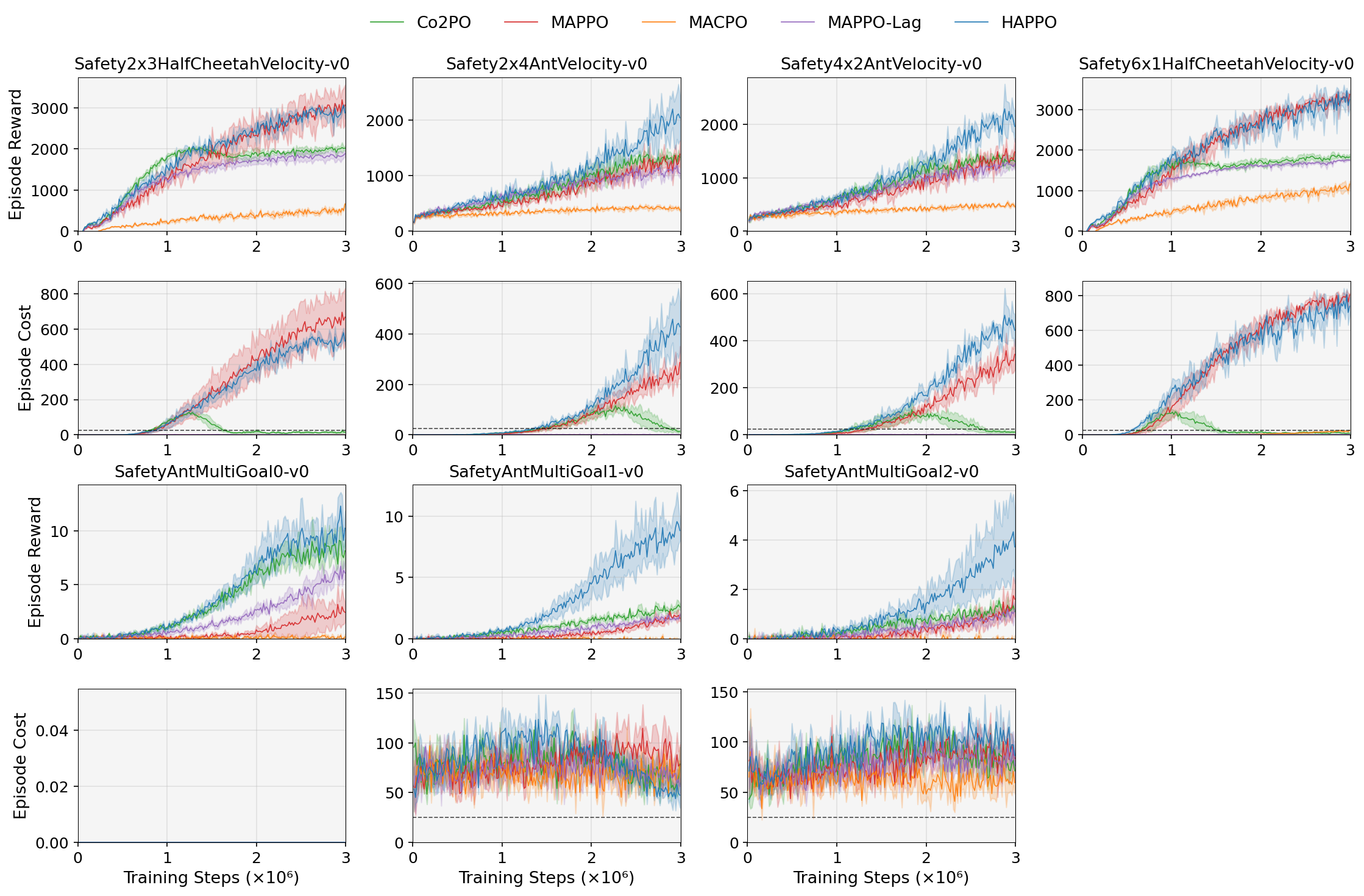}
  \caption{%
Results on seven \textsc{SafePO} cooperative safety benchmarks.
We report both evaluation return and cost.
  }
  \label{fig:co2po-results}
\end{figure*}

\section{Results}
We evaluate Co2PO on coordination-intensive environments under constrained multi-agent learning. Across all settings, Co2PO achieves higher feasible returns than leading baselines while maintaining evaluation-time cost compliance at convergence. To facilitate comparison across heterogeneous tasks, we aggregate feasible performance and evaluation-time cost by environment type. 
\label{sec:results}
\begin{table}[h!]
\centering
\caption{Mean feasible return across Velocity and Multi-goal environments, and mean final evaluation cost for Velocity tasks.}
\label{tab:feasible_by_env_type}
\begin{small}
\setlength{\tabcolsep}{4pt}
\begin{tabular}{lccc}
\toprule
\textbf{Method} 
& $\boldsymbol{R_{\text{feas}}}$ (Vel.) 
& $\boldsymbol{R_{\text{feas}}}$ (Multi.) 
& $\boldsymbol{C_{\text{final}}}$ (Vel.) \\
\midrule
MAPPO 
& $866$ 
& $3.38$ 
& $525.34$ \\
MAPPO-Lag 
& $1593$ 
& $7.09$ 
& $1.54$ \\
MACPO 
& $738$ 
& $0.69$ 
& $6.94$ \\
HAPPO 
& $1023$ 
& $13.00$ 
& $536.05$ \\
\midrule
\textbf{Co2PO} 
& $\boldsymbol{1709}$ 
& $\boldsymbol{10.00}$ 
& $\boldsymbol{12.30}$ \\
\bottomrule
\end{tabular}
\end{small}
\end{table}

\subsection{Feasible Performance Under Safety Constraints}
\label{sec:feasible}
Co2PO consistently identifies higher-quality feasible solutions throughout training. 
Across velocity-based environments, Co2PO achieves a mean feasible-return improvement of approximately $7\%$ relative to MAPPO-Lag, with a median improvement of $7\%$ (\cref{tab:feasible_by_env_type}). 
These gains are consistent across all velocity tasks, ranging from $6.5\%$ to $7.9\%$ depending on the environment.

In the coordination-intensive multi-goal environment with a feasible region, Co2PO achieves an absolute feasible-return improvement of $2.91$ over MAPPO-Lag (see Appendix~\ref{app:results} for all per-environment statistics). Although the percentage gains are inflated by small baseline values, the absolute improvement confirms Co2PO’s effectiveness in coordination-heavy regimes.

Co2PO demonstrates consistent deployment readiness on the Velocity family: final evaluation costs stabilize across seeds, and cost-compliant checkpoints are reliably reached while maintaining strong returns. 
In all MultiGoal settings, no method reaches the cost budget within the training horizon, so we focus on return--cost tradeoffs rather than feasible return (\cref{sec:ablations}).

\subsection{Learning Dynamics and Exploration--Safety Tradeoff}
\begin{table}[h!]
  \caption{Mean early return and peak episodic cost for velocity environments.}
  \label{tab:exploration_safety_tradeoff}
  \centering
  \begin{small}
  \setlength{\tabcolsep}{6pt}
  \begin{tabular}{l c c}
    \toprule
    \textbf{Method} & $\boldsymbol{C_{\text{peak}}}$ (Vel.) & $\boldsymbol{R_{\text{early}}}$ (Vel.) \\
    \midrule
    MAPPO & $855$ & $1023$ \\
    MAPPO-Lag & $2$ & $881$ \\
    MACPO & $79$ & $622$ \\
    HAPPO & $792$ & $1124$ \\
    \midrule
    \textbf{Co2PO} & $\boldsymbol{231}$ & $\boldsymbol{1221}$ \\
    \bottomrule
  \end{tabular}
  \end{small}
  \vskip -0.08in
\end{table}

Peak episodic cost $C_{\text{peak}}$ characterizes worst-case constraint violations during training and highlights differences in exploration behavior. 
Co2PO permits bounded early violations, incurring higher peak cost than conservative constrained baselines while remaining substantially below unconstrained methods (\cref{tab:exploration_safety_tradeoff}). 
This behavior is confined to early training and reflects an intentional relaxation of constraint enforcement to promote exploration.

This relaxation enables faster reward acquisition. 
Over the first $10^6$ environment steps, Co2PO achieves higher early evaluation return than MAPPO-Lag and MACPO, indicating more rapid learning progress under partial observability. 
In contrast, methods that aggressively suppress early violations exhibit lower peak cost but slower reward growth and ultimately converge to lower feasible return (\cref{sec:feasible}). 
By tolerating transient violations while enforcing constraints at convergence, Co2PO accelerates learning toward higher-quality, deployable policies.

\subsection{Deployment Safety and Stability}

We evaluate deployment safety using final evaluation cost $C_{\text{final}}$ and violation statistics at convergence. 
Across velocity environments, Co2PO exhibits stable and bounded cost behavior at evaluation time, with low variation across tasks (coefficient of variation, CV $\approx 0.20$), indicating predictable deployment performance. This stability is comparable to that of constrained baselines (MAPPO-Lag CV $\approx 0.29$, MACPO CV $\approx 1.53$), showing that Co2PO’s constraint satisfaction generalizes consistently across environments.

Safety constraint satisfaction at convergence is consistent, with a mean standard deviation of approximately $8\%$ across velocity environments. 
Final costs remain within a feasible range (mean $C_{\text{final}} = 12.30$, range $8.85$--$15.17$), indicating reliable constraint enforcement without erratic cost dynamics. The low cost variance, bounded violations, and reliable feasibility attainment indicate that Co2PO’s constraint satisfaction at convergence is consistent and suitable for deployment, rather than brittle or noise-sensitive.

\subsection{Environment-Dependent Behavioral Patterns}

Co2PO’s performance varies systematically with coordination demands. The largest improvements occur in environments requiring explicit multi-agent coordination, such as \textsc{AntVel}, where Co2PO achieves substantially higher feasible return and more reliable constraint satisfaction than constrained baselines ($R_{\text{feas}} \approx 1{,}450$ vs.\ $\approx 1{,}350$ for MAPPO-Lag across \textsc{AntVel} environments; Appendix~\ref{app:results}). In these settings, reactive and unconstrained baselines frequently fail to converge to compliant solutions within the training horizon: MAPPO and HAPPO do not achieve feasibility in all of the environments, including \textsc{HalfCheetahVel} variants. 

In contrast, environments with simpler dynamics and weaker inter-agent coupling, such as single-direction \textsc{HalfCheetahVel} variants, exhibit more modest but consistent gains ($R_{\text{feas}} \approx 1{,}950$ vs.\ $\approx 1{,}850$), while maintaining comparable stability at convergence. A similar trend appears in multi-goal environments, where Co2PO yields large relative improvements (approximately $40\%$ over MAPPO-Lag; Appendix~\ref{app:results}). Overall, Co2PO’s benefits are environment-dependent rather than uniform, scaling with coordination demands while generalizing across diverse tasks.

\subsection{Comparative Safety–Performance Analysis}

The evaluated baselines exhibit distinct safety--performance tradeoffs. 
Unconstrained methods such as HAPPO and MAPPO achieve high returns but incur substantial costs and elevated violation rates. 
MAPPO-Lag reliably enforces constraints but converges to overly conservative policies, resulting in markedly reduced feasible return. 
MACPO occupies an intermediate regime, mitigating some violations while allowing limited exploration, but consistently underperforms Co2PO across all evaluated environments.

Co2PO differs by enabling proactive constraint enforcement. 
Through hazard prediction and selective inter-agent communication, Co2PO permits controlled early exploration without sacrificing constraint satisfaction at convergence.
This yields a consistently improved safety--performance tradeoff, combining both higher feasible return and reliable deployment-time safety.

Overall, these results suggest that coupling selective coordination with predicted risk provides a practical mechanism for balancing exploration and constraint satisfaction. Importantly, Co2PO’s gains do not stem from constraint enforcement alone, but from improved coordination efficiency that allows agents to reach higher feasible returns.

\section{Ablations}
\label{sec:ablations}

\begin{figure*}[h!]
  \centering
  \includegraphics[width=0.99\textwidth]{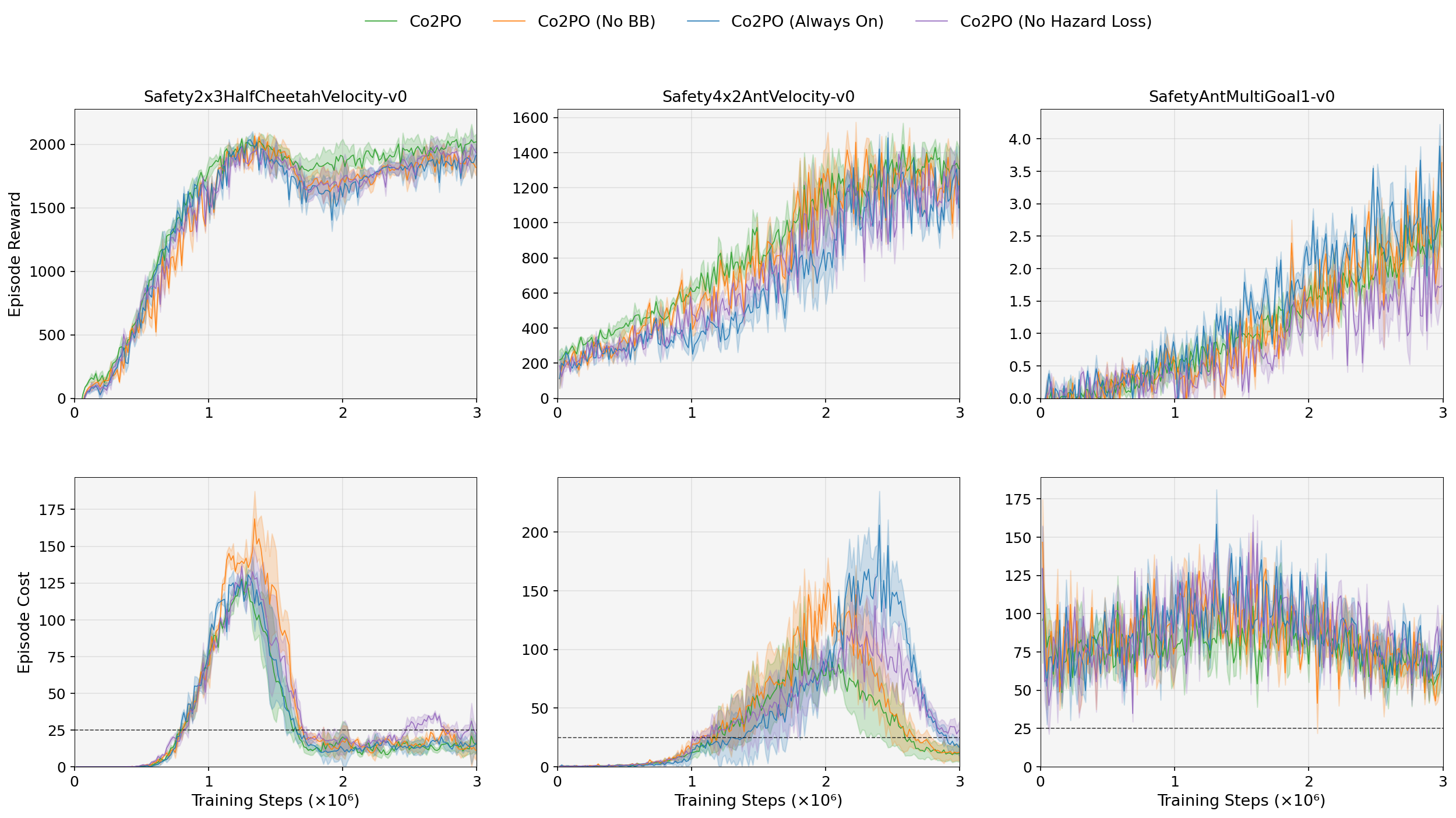}
  \caption{%
    Ablations on three environments. We report both evaluation return and cost.
  }
  \label{fig:co2po-ablations}
\end{figure*}

\begin{table*}[h!]
\centering
\small
\setlength{\tabcolsep}{4pt}
\caption{\textbf{Ablation summary.} For Velocity tasks we report \emph{feasible return} and the final evaluation cost $C_{\mathrm{final}}$ at the last checkpoint.
For \textsc{AntMultiGoal}, no method reaches the budget within the plotted horizon, so we report final return $R_{\mathrm{final}}$ instead.}
\label{tab:ablation_co2po}
\begin{tabular}{lcccccc}
\toprule
& \multicolumn{2}{c}{HalfCheetahVel} & \multicolumn{2}{c}{AntVel} & \multicolumn{2}{c}{AntMultiGoal} \\
\cmidrule(lr){2-3}\cmidrule(lr){4-5}\cmidrule(lr){6-7}
\textbf{Method} &
$R_{\mathrm{feas}}\uparrow$ & $C_{\mathrm{final}}\downarrow$ &
$R_{\mathrm{feas}}\uparrow$ & $C_{\mathrm{final}}\downarrow$ &
$R_{\mathrm{final}}\uparrow$ & $C_{\mathrm{final}}\downarrow$ \\
\midrule
\textsc{Co2PO} & 2054 $\pm$ 70 & 15.2 $\pm$ 7.9 & 1468 $\pm$ 43 & 11.3 $\pm$ 8.8 & 2.59 $\pm$ 0.32 & 78.8 $\pm$ 20.8 \\
\textsc{Co2PO} w/o Blackboard & 1907 $\pm$ 65 & 12.3 $\pm$ 6.4 & 1365 $\pm$ 40 & 13.7 $\pm$ 10.6 & 3.46 $\pm$ 0.42 & 84.4 $\pm$ 22.3 \\
\textsc{Co2PO} (Always-Write) & 1915 $\pm$ 65 & 16.6 $\pm$ 8.6 & 1284 $\pm$ 38 & 14.9 $\pm$ 11.6 & 3.07 $\pm$ 0.38 & 67.0 $\pm$ 17.7 \\
\textsc{Co2PO} w/o Hazard Loss & 1936 $\pm$ 66 & 23.1 $\pm$ 12.0 & 566 $\pm$ 17 & 28.2 $\pm$ 21.9 & 1.98 $\pm$ 0.24 & 74.6 $\pm$ 19.7 \\
\bottomrule
\end{tabular}
\end{table*}

We ablate Co2PO's coordination components to isolate which mechanisms drive improvements in feasible performance.
We focus on three representative environments that span fast locomotion with interaction-driven hazards (\textsc{HalfCheetahVel}, \textsc{AntVel}) and a harder navigation-and-coordination setting (\textsc{AntMultiGoal}).
All ablations keep the same training budget and settings as the full method. We additionally study the effect of the hazard lookahead horizon in Appendix~\ref{app:lookahead}.

\textbf{Takeaway.}
\cref{fig:co2po-ablations,tab:ablation_co2po} shows that removing \emph{either} (i) shared-memory coordination (\emph{w/o Blackboard}) or (ii) selective gating (\emph{Always-Write}) reduces feasible performance on the Velocity tasks.
The most pronounced degradation occurs when removing hazard supervision (\emph{w/o Hazard Loss}), which substantially worsens both feasibility and performance in \textsc{AntVel}.

\subsection{Shared Memory Is Beneficial for Feasible Performance}
\label{sec:ablate-blackboard}

\textbf{w/o Blackboard} removes inter-agent shared context, while keeping the rest of training unchanged.
On \textsc{HalfCheetahVel} and \textsc{AntVel}, this consistently lowers feasible return relative to Co2PO (e.g., $2054\!\rightarrow\!1907$ on \textsc{HalfCheetahVel} and $1468\!\rightarrow\!1365$ on \textsc{AntVel}), indicating that the blackboard context is contributing useful coordination information beyond what is recoverable from local observations alone.
Final evaluation costs remain below budget in these Velocity settings, suggesting the main effect of removing the blackboard is reduced \emph{feasible performance} rather than purely increased constraint violations.

In \textsc{AntMultiGoal}, none of the compared methods reaches the budget within the plotted horizon; we therefore report final return instead of feasible return.
Here, removing the blackboard increases $R_{\mathrm{final}}$ but also increases $C_{\mathrm{final}}$ (Table~\ref{tab:ablation_co2po}), highlighting that this environment exhibits a different trade-off regime where constraint satisfaction is not yet achieved by any method in the training budget we consider.

\subsection{Selective Writes Matter}
\label{sec:ablate-always-write}

\textbf{Always-Write} disables sparsity by forcing all agents to write every timestep, while keeping the same message format and read mechanism.
Across the Velocity tasks, Always-Write underperforms the full method in feasible return (e.g., $2054\!\rightarrow\!1915$ on \textsc{HalfCheetahVel} and $1468\!\rightarrow\!1284$ on \textsc{AntVel}), with final costs that remain broadly comparable.
This supports the practical motivation for selective coordination: indiscriminate writing does not reliably improve feasible performance under a fixed training budget, and can introduce additional noise into the retrieved context.

On \textsc{AntMultiGoal}, Always-Write yields a lower final cost than the full method but does not resolve infeasibility; all variants remain above the budget.
We treat this setting as a stress test as none of the baselines reach compliance with regard to the set cost target.

\subsection{Hazard Supervision Is Important for Feasibility}
\label{sec:ablate-hazard}

\textbf{w/o Hazard Loss} removes explicit supervision of the hazard predictor while leaving the coordination pathway intact.
This ablation has the most severe impact on \textsc{AntVel}: feasible return collapses ($1468\!\rightarrow\!566$) and final cost rises above the budget ($11.3\!\rightarrow\!28.2$).
The learning curves in \cref{fig:co2po-ablations} are consistent with the intuition that without direct hazard learning signal, the write trigger becomes less aligned with safety-relevant events, reducing the effectiveness of selective coordination and degrading constraint satisfaction.

On \textsc{HalfCheetahVel}, the same ablation increases final cost while still remaining under budget on average (Table~\ref{tab:ablation_co2po}), and yields a modest drop in feasible return relative to the full method.
On \textsc{AntMultiGoal}, removing hazard supervision decreases final return without materially changing the overall infeasibility regime.

\section{Related Work}
\label{sec:related}

\textbf{Constrained Reinforcement Learning.}
Constrained reinforcement learning (CRL) addresses policy optimization under explicit safety, resource, or performance constraints, most commonly via Lagrangian relaxation that embeds costs into the objective through adaptive penalty coefficients \cite{zhang2019marlsurvey}.
Single-agent methods such as CPO \cite{achiam2017constrainedpolicyoptimization}, PPO with Lagrangian relaxation \cite{ray2019safexp}, and FOCOPS \cite{zhang2020orderconstrainedoptimizationpolicy} enforce constraints in expectation, but are often sensitive to penalty tuning and rely on reactive updates that correct violations only after they occur, which can slow learning or induce conservative behavior \cite{stooke2020responsive}.
Extensions to multi-agent settings adopt similar principles, employing centralized critics or shared multipliers (e.g., MAPPO-Lagrangian and MACPO) to enforce global constraints.
While effective at reducing violations, these approaches largely rely on reactive, penalty-based constraint handling and can suppress exploration in coordination-intensive environments.

\textbf{Safe Multi-Agent Reinforcement Learning.} Safe multi-agent reinforcement learning (Safe MARL) further considers cooperative settings where safety violations arise from collective interactions rather than isolated actions. Most safe MARL methods operate under the centralized training with decentralized execution (CTDE) paradigm and augment standard MARL backbones such as MAPPO with global cost critics or shared constraint signals \cite{hernandezleal2019surveylearningmultiagentenvironments, foerster2024counterfactualmultiagentpolicygradients}, rather than value-factorization approaches such as QMIX \cite{sunehag2017valuedecompositionnetworkscooperativemultiagent, rashid2018qmixmonotonicvaluefunction}. However, safety is typically enforced through aggregated penalties that provide limited guidance for resolving coordination-induced violations, leading to slow learning dynamics or overly conservative policies \cite{gu2023macpo}. Prior work in hierarchical MARL demonstrates that explicit coordination improves task performance, suggesting that similar mechanisms may be beneficial for safety-constrained multi-agent settings.

\textbf{Communication and Memory in MARL.} Communication has been widely studied as a means of improving coordination in MARL, particularly under partial observability, through message passing or graph-based information sharing \cite{sukhbaatar2016learningmultiagentcommunicationbackpropagation, kim2019schednet, das2020tarmactargetedmultiagentcommunication, jiang2020graphconvolutionalreinforcementlearning}.
Existing approaches primarily use communication to enhance task performance and learning stability 
\cite{peng2021copo,calzolari2026safeheterogeneousmultiagentrl}, without explicitly targeting safety constraints. Some methods combine communication with safety mechanisms by coupling MAPPO-style training with execution-time safety filters or shields \cite{alshiekh2017safereinforcementlearningshielding}, while others incorporate shared cost signals during training to enforce constraints.
In both cases, safety is handled reactively via centralized penalties or post-hoc intervention, and communication is rarely conditioned on anticipated risk or collective hazard.

Co2PO departs from prior work by coupling communication with safety objectives as part of the learning process, rather than engineering safety as an external rule. In addition to Lagrangian constraint penalties, Co2PO leverages hazard prediction to selectively trigger communication and coordination when safety risks are anticipated. Communication is neither constant nor purely performance-driven, but explicitly conditioned on predicted constraint violations, allowing more targeted and efficient safety coordination.

\section{Discussion}
\label{sec:conclusion}

\textbf{Why Co2PO helps.}
Our central hypothesis is that many safety failures in cooperative MARL are \emph{coordination-induced}: violations arise from mismatched timing, intent, or mutual awareness under partial observability. Co2PO addresses this by turning communication into a \emph{risk-triggered control signal}. Agents only broadcast when their hazard predictor anticipates near-term risk, and the blackboard carries compact, safety-relevant semantics (state summary, intent, yield, hazard probability) that other agents can immediately condition on. This differs from purely reactive constraint handling, where penalties increase \emph{after} violations and can suppress exploration globally \cite{brunke2022safelearning, gu2023macpo}.

\textbf{Empirical takeaways.}
Across seven \textsc{SafePO} benchmarks, Co2PO consistently finds higher-quality feasible checkpoints than strong constrained baselines, especially on the Velocity family where interaction-driven hazards are common (\cref{fig:co2po-results}). The learning curves also reflect an explicit exploration--safety strategy: Co2PO tolerates bounded early violations (higher $C_{\text{peak}}$) to acquire reward faster, then improves constraint satisfaction and stabilizes by deployment (\cref{sec:results}). Ablations isolate the key mechanisms, justifying the role of the blackboard, selective writes, and hazard supervision (\cref{fig:co2po-ablations,tab:ablation_co2po}). In the appendix, varying the lookahead horizon further supports the role of proactive labeling as nonzero horizons can improve the return--cost tradeoff relative to $H\!=\!0$ (\cref{app:lookahead}).

\textbf{Limitations and future work.}
First, Co2PO is designed to target \emph{deployment-safe policies under a fixed training budget}, rather than zero-violation training-time safety \cite{alshiekh2017safereinforcementlearningshielding, berkenkamp2017safemodelbasedreinforcementlearning}: peak violations can be larger than conservative baselines, which may be unacceptable in some real systems.
Second, our hazard labels rely on dense per-step costs and a hand-chosen threshold $\delta$ (and horizon $H$), which may not be available or stable across domains; learning hazard signals from sparse events, uncertainty-aware predictors, or model-based rollouts is a natural extension \cite{yang2020meanfieldmultiagentreinforcement, calzolari2026safeheterogeneousmultiagentrl}. Third, the Multi-Goal regime remains challenging under our fixed budget, suggesting we need training signals that better reflect delayed coordination effects (e.g., when early yielding prevents later conflicts). Finally, we have not yet stressed scalability: blackboard retrieval and message semantics may need compression, structured routing, or graph-based selection as agent counts grow \cite{goeckner2024graphbasedmar}. Together, these directions would broaden Co2PO’s applicability and improve robustness of risk-triggered coordination across domains, horizons, and agent scales.

\clearpage


\section*{Impact Statement}
This work advances methods for learning policies under safety constraints in multi-agent environments. Improved constrained MARL can reduce unsafe behavior in high-stakes domains where multiple decision-makers interact. Potential risks include misuse in adversarial settings or deployment without adequate monitoring; we encourage careful evaluation, transparent reporting of training-time violations, and robust safety testing prior to real-world use.

\bibliographystyle{icml2026}
\bibliography{citations}

@misc{achiam2017constrainedpolicyoptimization,
  title        = {Constrained Policy Optimization},
  author       = {Joshua Achiam and David Held and Aviv Tamar and Pieter Abbeel},
  year         = {2017},
  archivePrefix= {arXiv},
  primaryClass = {cs.LG},
  note         = {arXiv:1705.10528}
}

@misc{ray2019safexp,
  author = {Alex Ray and Joshua Achiam and Dario Amodei},
  title  = {Benchmarking Safe Exploration in Deep Reinforcement Learning},
  year   = {2019},
  note   = {OpenAI Technical Report}
}

@misc{tessler2018rcpo,
  author    = {Chen Tessler and Daniel J. Mankowitz and Shie Mannor},
  title     = {Reward Constrained Policy Optimization},
  year      = {2018},
  note      = {arXiv:1805.11074}
}

@misc{stooke2020responsive,
  author    = {Adam Stooke and Pieter Abbeel},
  title     = {Responsive Safety in Reinforcement Learning by PID Lagrangian Methods},
  year      = {2020},
  note      = {arXiv:2007.03964}
}

@misc{yu2022surprisingeffectivenessppocooperative,
  title        = {The Surprising Effectiveness of PPO in Cooperative, Multi-Agent Games},
  author       = {Chao Yu and Akash Velu and Eugene Vinitsky and Jiaxuan Gao and Yu Wang and Alexandre Bayen and Yi Wu},
  year         = {2022},
  archivePrefix= {arXiv},
  primaryClass = {cs.LG},
  note         = {arXiv:2103.01955}
}

@misc{kuba2022happo,
  author    = {Jakub Kuba and Yuchen Jiang and Wei Wei and Jianye Hao},
  title     = {Trust Region Multi-Agent Policy Optimization},
  year      = {2022},
  note      = {arXiv:2109.11251}
}

@misc{lowe2020multiagentactorcriticmixedcooperativecompetitive,
  title        = {Multi-Agent Actor-Critic for Mixed Cooperative-Competitive Environments},
  author       = {Ryan Lowe and Yi Wu and Aviv Tamar and Jean Harb and Pieter Abbeel and Igor Mordatch},
  year         = {2020},
  archivePrefix= {arXiv},
  primaryClass = {cs.LG},
  note         = {arXiv:1706.02275}
}

@misc{rashid2018qmixmonotonicvaluefunction,
  title        = {QMIX: Monotonic Value Function Factorisation for Deep Multi-Agent Reinforcement Learning},
  author       = {Tabish Rashid and Mikayel Samvelyan and Christian Schroeder de Witt and Gregory Farquhar and Jakob Foerster and Shimon Whiteson},
  year         = {2018},
  archivePrefix= {arXiv},
  primaryClass = {cs.LG},
  note         = {arXiv:1803.11485}
}

@misc{foerster2024counterfactualmultiagentpolicygradients,
  title        = {Counterfactual Multi-Agent Policy Gradients},
  author       = {Jakob Foerster and Gregory Farquhar and Triantafyllos Afouras and Nantas Nardelli and Shimon Whiteson},
  year         = {2024},
  archivePrefix= {arXiv},
  primaryClass = {cs.AI},
  note         = {arXiv:1705.08926}
}

@misc{gu2023macpo,
  author    = {Zheng Gu and Qiang Fu and Hao Dong},
  title     = {Safe Multi-Agent Reinforcement Learning for Multi-Robot Real-Time Systems},
  year      = {2023},
  note      = {arXiv:2110.02793}
}

@misc{das2020tarmactargetedmultiagentcommunication,
  title        = {TarMAC: Targeted Multi-Agent Communication},
  author       = {Abhishek Das and Théophile Gervet and Joshua Romoff and Dhruv Batra and Devi Parikh and Michael Rabbat and Joelle Pineau},
  year         = {2020},
  archivePrefix= {arXiv},
  primaryClass = {cs.LG},
  note         = {arXiv:1810.11187}
}

@misc{zhang2019marlsurvey,
  author    = {Kaiqing Zhang and Zhuoran Yang and Tamer Ba{\c{s}}ar},
  title     = {Multi-Agent Reinforcement Learning: A Selective Overview of Theories and Algorithms},
  year      = {2019},
  note      = {arXiv:1911.10635}
}

@misc{kim2019schednet,
  author    = {Jihwan Kim and Jinkyoo Park and Yoonho Kim and Sungwon Park},
  title     = {Learning to Schedule Communication in Multi-Agent Reinforcement Learning},
  year      = {2019},
  note      = {arXiv:1902.01554}
}

@misc{peng2021copo,
  author    = {Zhenghao Peng and Quan Vuong and Brandon Amos and others},
  title     = {Coordinated Policy Optimization},
  year      = {2021},
  note      = {arXiv:2110.13827}
}

@misc{schulman2017proximalpolicyoptimizationalgorithms,
  title        = {Proximal Policy Optimization Algorithms},
  author       = {John Schulman and Filip Wolski and Prafulla Dhariwal and Alec Radford and Oleg Klimov},
  year         = {2017},
  archivePrefix= {arXiv},
  primaryClass = {cs.LG},
  note         = {arXiv:1707.06347}
}

@misc{brunke2022safelearning,
  author    = {Lukas Brunke and others},
  title     = {Safe Learning in Robotics: From Guarantees to Implementations},
  year      = {2022},
  note      = {arXiv:2108.06266}
}

@misc{alshiekh2017safereinforcementlearningshielding,
  title        = {Safe Reinforcement Learning via Shielding},
  author       = {Mohammed Alshiekh and Roderick Bloem and R{\"u}diger Ehlers and Bettina K{\"o}nighofer and Scott Niekum and Ufuk Topcu},
  year         = {2017},
  archivePrefix= {arXiv},
  primaryClass = {cs.LO},
  note         = {arXiv:1708.08611}
}

@misc{calzolari2026safeheterogeneousmultiagentrl,
  title        = {Safe Heterogeneous Multi-Agent RL with Communication Regularization for Coordinated Target Acquisition},
  author       = {Gabriele Calzolari and Vidya Sumathy and Christoforos Kanellakis and George Nikolakopoulos},
  year         = {2026},
  archivePrefix= {arXiv},
  primaryClass = {cs.RO},
  note         = {arXiv:2601.08327}
}

@misc{sukhbaatar2016learningmultiagentcommunicationbackpropagation,
  title        = {Learning Multiagent Communication with Backpropagation},
  author       = {Sainbayar Sukhbaatar and Arthur Szlam and Rob Fergus},
  year         = {2016},
  archivePrefix= {arXiv},
  primaryClass = {cs.LG},
  note         = {arXiv:1605.07736}
}

@misc{ji2024safetygymnasiumunifiedsafereinforcement,
  title        = {Safety-Gymnasium: A Unified Safe Reinforcement Learning Benchmark},
  author       = {Jiaming Ji and Borong Zhang and Jiayi Zhou and Xuehai Pan and Weidong Huang and Ruiyang Sun and Yiran Geng and Yifan Zhong and Juntao Dai and Yaodong Yang},
  year         = {2024},
  archivePrefix= {arXiv},
  primaryClass = {cs.AI},
  note         = {arXiv:2310.12567}
}

@misc{zhang2020orderconstrainedoptimizationpolicy,
  title        = {First Order Constrained Optimization in Policy Space},
  author       = {Yiming Zhang and Quan Vuong and Keith W. Ross},
  year         = {2020},
  archivePrefix= {arXiv},
  primaryClass = {cs.LG},
  note         = {arXiv:2002.06506}
}

@misc{yang2020meanfieldmultiagentreinforcement,
  title        = {Mean Field Multi-Agent Reinforcement Learning},
  author       = {Yaodong Yang and Rui Luo and Minne Li and Ming Zhou and Weinan Zhang and Jun Wang},
  year         = {2020},
  archivePrefix= {arXiv},
  primaryClass = {cs.MA},
  note         = {arXiv:1802.05438}
}

@misc{sunehag2017valuedecompositionnetworkscooperativemultiagent,
  title        = {Value-Decomposition Networks for Cooperative Multi-Agent Learning},
  author       = {Peter Sunehag and Guy Lever and Audrunas Gruslys and Wojciech Marian Czarnecki and Vinicius Zambaldi and Max Jaderberg and Marc Lanctot and Nicolas Sonnerat and Joel Z. Leibo and Karl Tuyls and Thore Graepel},
  year         = {2017},
  archivePrefix= {arXiv},
  primaryClass = {cs.AI},
  note         = {arXiv:1706.05296}
}

@misc{jiang2020graphconvolutionalreinforcementlearning,
  title        = {Graph Convolutional Reinforcement Learning},
  author       = {Jiechuan Jiang and Chen Dun and Tiejun Huang and Zongqing Lu},
  year         = {2020},
  archivePrefix= {arXiv},
  primaryClass = {cs.LG},
  note         = {arXiv:1810.09202}
}

@misc{hernandezleal2019surveylearningmultiagentenvironments,
  title        = {A Survey of Learning in Multiagent Environments: Dealing with Non-Stationarity},
  author       = {Pablo Hernandez-Leal and Michael Kaisers and Tim Baarslag and Enrique Munoz de Cote},
  year         = {2019},
  archivePrefix= {arXiv},
  primaryClass = {cs.MA},
  note         = {arXiv:1707.09183}
}

@misc{berkenkamp2017safemodelbasedreinforcementlearning,
  title        = {Safe Model-based Reinforcement Learning with Stability Guarantees},
  author       = {Felix Berkenkamp and Matteo Turchetta and Angela P. Schoellig and Andreas Krause},
  year         = {2017},
  archivePrefix= {arXiv},
  primaryClass = {stat.ML},
  note         = {arXiv:1705.08551}
}

@misc{omidshafiei2015decentralizedcontrolpartiallyobservable,
  title        = {Decentralized Control of Partially Observable Markov Decision Processes using Belief Space Macro-actions},
  author       = {Shayegan Omidshafiei and Ali-akbar Agha-mohammadi and Christopher Amato and Jonathan P. How},
  year         = {2015},
  archivePrefix= {arXiv},
  primaryClass = {cs.MA},
  note         = {arXiv:1502.06030}
}

@INPROCEEDINGS{6386109,
  author={Todorov, Emanuel and Erez, Tom and Tassa, Yuval},
  booktitle={2012 IEEE/RSJ International Conference on Intelligent Robots and Systems}, 
  title={MuJoCo: A physics engine for model-based control}, 
  year={2012},
  volume={},
  number={},
  pages={5026-5033},
  doi={10.1109/IROS.2012.6386109}}

@misc{goeckner2024graphbasedmar,
  title        = {Graph Neural Network-based Multi-Agent Reinforcement Learning for Resilient Distributed Coordination of Multi-Robot Systems},
  author       = {Anthony Goeckner and Yueyuan Sui and Nicolas Martinet and Xinliang Li and Qi Zhu},
  year         = {2024},
  archivePrefix= {arXiv},
  primaryClass = {cs.MA},
  note         = {arXiv:2403.13093}
}

@inproceedings{littman1994markovgames,
  title     = {Markov Games as a Framework for Multi-Agent Reinforcement Learning},
  author    = {Michael L. Littman},
  booktitle = {Proceedings of the 11th International Conference on Machine Learning (ICML)},
  pages     = {157--163},
  year      = {1994}
}
\clearpage



\newpage
\appendix
\section{Complete Evaluation Metrics}
\label{app:results}

\begin{table}[h!]
\centering
\caption{Full evaluation metrics aggregated across runs.}
\label{tab:full_metrics}
\resizebox{\textwidth}{!}{
\begin{tabular}{llcccccc}
\toprule
\textbf{Environment} & \textbf{Method} & $\boldsymbol{R_{\text{final}}}$ & $\boldsymbol{R_{\text{feas}}}$ & $\boldsymbol{C_{\text{final}}}$ & $\boldsymbol{C_{\text{peak}}}$ & \textbf{Violation rate} & \textbf{Time-to-feasible} \\
\midrule
\multirow{5 }{* }{}{Safety2x3HalfCheetahVelocity} & MAPPO & $3068 \pm 491$ & $1094 \pm 43$ & $667.17 \pm 164.13$ & $712$ & $0.71 \pm 0.02$ & -- \\
 & MAPPO-Lag & $1882 \pm 74$ & $1904 \pm 86$ & $1.10 \pm 0.51$ & $3$ & $0.00 \pm 0.00$ & $16000$ \\
 & MACPO & $657 \pm 22$ & $663 \pm 16$ & $0.48 \pm 0.21$ & $0$ & $0.00 \pm 0.00$ & $16000$ \\
 & HAPPO & $2878 \pm 143$ & $1362 \pm 95$ & $532.04 \pm 21.41$ & $583$ & $0.72 \pm 0.02$ & -- \\
 & \textbf{Co2PO} & $\mathbf{2018 \pm 61}$ & $\mathbf{2054 \pm 57}$ & $\mathbf{15.17 \pm 6.46}$ & $\mathbf{133}$ & $\mathbf{0.30 \pm 0.01}$ & $\mathbf{1704000}$ \\
\midrule
\multirow{5 }{* }{}{Safety2x4AntVelocity} & MAPPO & $1386 \pm 242$ & $704 \pm 37$ & $286.34 \pm 64.61$ & $305$ & $0.50 \pm 0.04$ & -- \\
 & MAPPO-Lag & $1038 \pm 89$ & $1315 \pm 51$ & $1.39 \pm 0.52$ & $3$ & $0.00 \pm 0.00$ & $16000$ \\
 & MACPO & $378 \pm 29$ & $493 \pm 22$ & $0.53 \pm 0.28$ & $1$ & $0.00 \pm 0.00$ & $16000$ \\
 & HAPPO & $2041 \pm 405$ & $933 \pm 60$ & $427.80 \pm 99.41$ & $494$ & $0.54 \pm 0.02$ & -- \\
 & \textbf{Co2PO} & $\mathbf{1330 \pm 41}$ & $\mathbf{1418 \pm 42}$ & $\mathbf{13.82 \pm 4.80}$ & $\mathbf{133}$ & $\mathbf{0.45 \pm 0.04}$ & $\mathbf{2810667}$ \\
\midrule
\multirow{5 }{* }{}{Safety4x2AntVelocity} & MAPPO & $1486 \pm 254$ & $698 \pm 38$ & $341.95 \pm 65.31$ & $383$ & $0.55 \pm 0.03$ & -- \\
 & MAPPO-Lag & $1253 \pm 103$ & $1374 \pm 92$ & $1.47 \pm 0.64$ & $4$ & $0.00 \pm 0.00$ & $16000$ \\
 & MACPO & $459 \pm 20$ & $559 \pm 22$ & $1.47 \pm 0.66$ & $2$ & $0.00 \pm 0.00$ & $16000$ \\
 & HAPPO & $1966 \pm 290$ & $749 \pm 42$ & $444.63 \pm 67.31$ & $570$ & $0.61 \pm 0.02$ & -- \\
 & \textbf{Co2PO} & $\mathbf{1308 \pm 121}$ & $\mathbf{1468 \pm 35}$ & $\mathbf{11.34 \pm 7.19}$ & $\mathbf{126}$ & $\mathbf{0.42 \pm 0.03}$ & $\mathbf{2528000}$ \\
\midrule
\multirow{5 }{* }{}{Safety6x1HalfCheetahVelocity} & MAPPO & $3381 \pm 4$ & $968 \pm 47$ & $805.90 \pm 9.45$ & $826$ & $0.76 \pm 0.02$ & -- \\
 & MAPPO-Lag & $1754 \pm 15$ & $1778 \pm 16$ & $2.19 \pm 0.98$ & $4$ & $0.00 \pm 0.00$ & $16000$ \\
 & MACPO & $1159 \pm 70$ & $1236 \pm 17$ & $25.29 \pm 2.43$ & $27$ & $0.01 \pm 0.00$ & $2912000$ \\
 & HAPPO & $3249 \pm 238$ & $1047 \pm 35$ & $739.71 \pm 51.41$ & $838$ & $0.82 \pm 0.01$ & -- \\
 & \textbf{Co2PO} & $\mathbf{1829 \pm 79}$ & $\mathbf{1894 \pm 55}$ & $\mathbf{8.85 \pm 4.65}$ & $\mathbf{154}$ & $\mathbf{0.29 \pm 0.04}$ & $\mathbf{1614667}$ \\
\midrule
\multirow{5 }{* }{}{SafetyAntMultiGoal-0} & MAPPO & $2.44 \pm 0.93$ & $3.38 \pm 1.73$ & $0.00 \pm 0.00$ & $0$ & $0.00 \pm 0.00$ & $16000$ \\
 & MAPPO-Lag & $6.08 \pm 0.63$ & $7.09 \pm 0.34$ & $0.00 \pm 0.00$ & $0$ & $0.00 \pm 0.00$ & $16000$ \\
 & MACPO & $0.14 \pm 0.29$ & $0.69 \pm 0.05$ & $0.00 \pm 0.00$ & $0$ & $0.00 \pm 0.00$ & $16000$ \\
 & HAPPO & $10 \pm 1$ & $13 \pm 1$ & $0.00 \pm 0.00$ & $0$ & $0.00 \pm 0.00$ & $16000$ \\
 & \textbf{Co2PO} & $\mathbf{8.13 \pm 1.05}$ & $\mathbf{10 \pm 1}$ & $\mathbf{0.00 \pm 0.00}$ & $\mathbf{0}$ & $\mathbf{0.00 \pm 0.00}$ & $\mathbf{16000}$ \\
\midrule
\multirow{5 }{* }{}{SafetyAntMultiGoal-1} & MAPPO & $1.66 \pm 0.24$ & -- & $78.21 \pm 20.21$ & $137$ & $1.00 \pm 0.00$ & -- \\
 & MAPPO-Lag & $1.93 \pm 0.15$ & -- & $56.30 \pm 7.64$ & $116$ & $1.00 \pm 0.00$ & -- \\
 & MACPO & $-0.12 \pm 0.05$ & -- & $76.46 \pm 5.70$ & $121$ & $1.00 \pm 0.00$ & -- \\
 & HAPPO & $8.84 \pm 0.85$ & -- & $47.38 \pm 12.59$ & $149$ & $1.00 \pm 0.00$ & -- \\
 & \textbf{Co2PO} & $\mathbf{2.59 \pm 0.26}$ & -- & $\mathbf{78.77 \pm 17.02}$ & $\mathbf{135}$ & $\mathbf{1.00 \pm 0.00}$ & -- \\
\midrule
\multirow{5 }{* }{}{SafetyAntMultiGoal-2} & MAPPO & $1.60 \pm 0.72$ & -- & $83.46 \pm 2.57$ & $124$ & $1.00 \pm 0.00$ & -- \\
 & MAPPO-Lag & $1.17 \pm 0.21$ & -- & $76.27 \pm 16.08$ & $127$ & $1.00 \pm 0.00$ & -- \\
 & MACPO & $-0.07 \pm 0.20$ & -- & $59.42 \pm 11.00$ & $122$ & $1.00 \pm 0.00$ & -- \\
 & HAPPO & $3.72 \pm 1.48$ & -- & $86.57 \pm 2.35$ & $151$ & $1.00 \pm 0.00$ & -- \\
 & \textbf{Co2PO} & $\mathbf{1.11 \pm 0.03}$ & -- & $\mathbf{71.27 \pm 0.45}$ & $\mathbf{133}$ & $\mathbf{1.00 \pm 0.00}$ & -- \\
\bottomrule
\end{tabular}
}
\end{table}


\onecolumn


\section{Effect of Hazard Lookahead Horizon}
\label{app:lookahead}

We study how the hazard lookahead horizon $H$ used to construct the proactive label $h_t^i$ in \cref{eq:lookahead-hazard} affects final performance. We compare Co2PO variants with $H\!\in\!\{0,3,5,8\}$, using the same training budget ($3{,}000{,}000$ steps) and evaluation protocol. Table~\ref{tab:lookahead} summarizes final-timestep statistics.

Overall, adding a nonzero lookahead yields \emph{consistent} improvements on the coordination-intensive Velocity family: all $H\!>\!0$ variants reduce average final cost relative to $H\!=\!0$ (while slightly increasing average final reward). The clearest improvement appears at the highest horizon ($H\!=\!8$). We also observe that higher horizons increase the number of environments that are cost-compliant at the final checkpoint (from $4/7$ to $5/7$). Longer lookahead ($H\!=\!8$) remains better than $H\!=\!0$ on Velocity costs, but does not further improve the harder MultiGoal settings in our runs (where final costs remain above budget for most methods).

\begin{table}[h!]
\centering
\caption{\textbf{Hazard lookahead horizon $H$ (final-timestep).} We report averages over the four Velocity environments (\textsc{HalfCheetahVel} and \textsc{AntVel}) and the provided averages over all $7$ environments. ``Feasible envs'' counts how many environments satisfy $C_{\text{final}}\le d$ at the final checkpoint (budget $d=25$).}
\label{tab:lookahead}
\small
\setlength{\tabcolsep}{6pt}
\renewcommand{\arraystretch}{1.08}
\begin{tabular}{lcccccc}
\toprule
\textbf{Method} & $\boldsymbol{H}$ &
$\overline{R}_{\text{final}}$ (Vel)$\uparrow$ &
$\overline{C}_{\text{final}}$ (Vel)$\downarrow$ &
\textbf{Feasible envs} $\uparrow$ \\
\midrule
\textsc{Co2PO} & 0 &
1607.79 & 19.34 &
$4/7$ \\
\textsc{Co2PO} & 3 &
1621.88 & 12.30 &
$4/7$ \\
\textsc{Co2PO} & 5 &
1639.69 & 16.32 &

$5/7$ \\
\textsc{Co2PO} & 8 &
1644.18 & 9.42 &

$5/7$ \\
\bottomrule
\end{tabular}
\vspace{-0.5em}
\end{table}

\paragraph{Takeaway.}
These results are consistent with the intuition behind proactive labeling: predicting hazards a few steps ahead provides a cleaner trigger for coordination, improving the safety--performance tradeoff on interaction-driven Velocity tasks.

\clearpage
\section{Method Diagram}

\begin{figure*}[h!]
  \centering
  \includegraphics[width=0.87\textwidth]{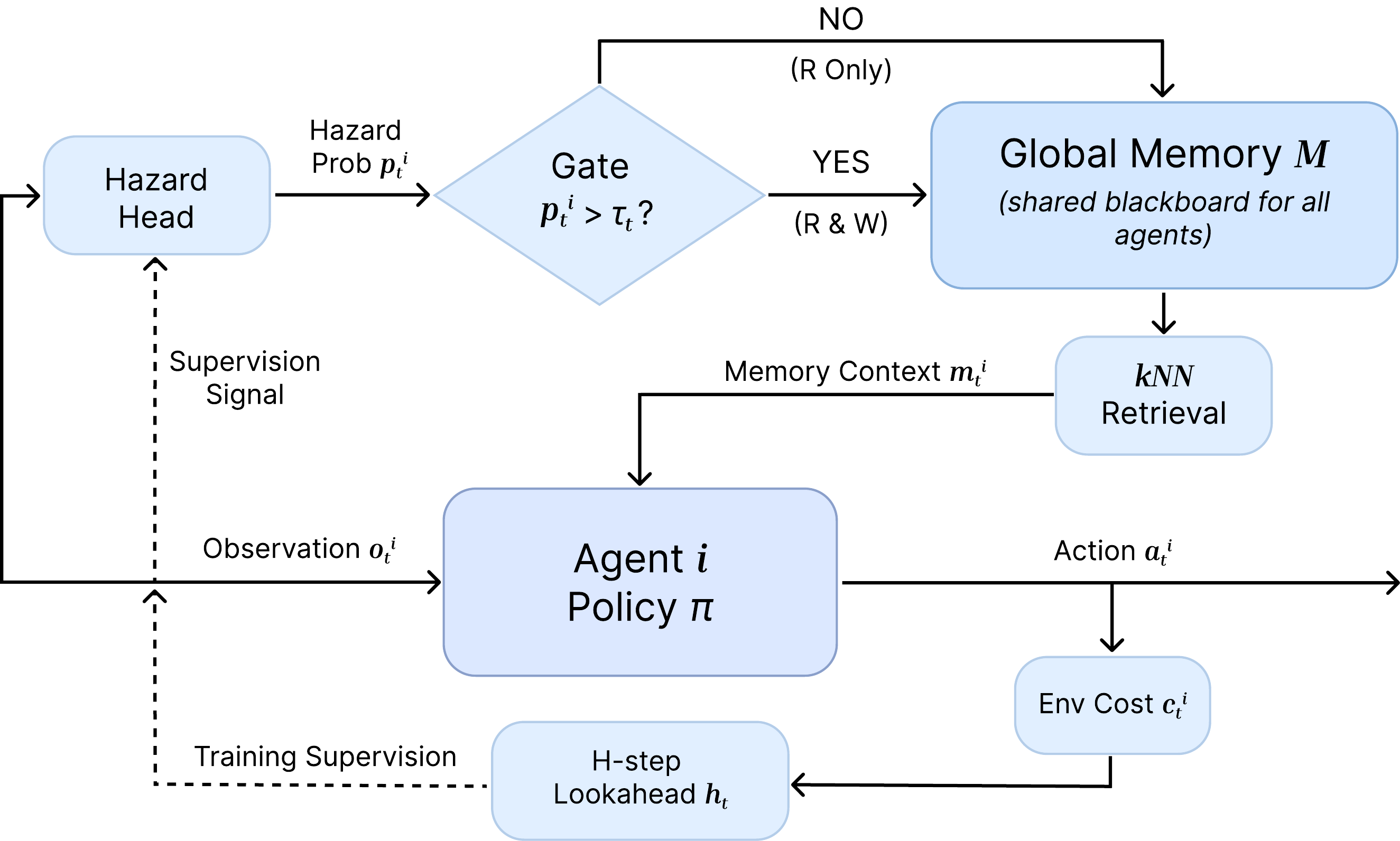}
  \caption{%
    Co2PO Framework Diagram
  }
  \label{fig:co2po-ablations}
\end{figure*}

\clearpage
\section{Detailed Hyperparameters}
\label{app:hparams}

\vfill
\begin{table}[h!]
\caption{\textbf{Key method hyperparameters (1/3).} Common run settings are in Table~\ref{tab:common-run-settings}.}
\label{tab:hparams-page1}
\centering
\begin{subtable}[t]{0.48\textwidth}
\caption{\textsc{Co2PO}}
\label{tab:hparams-co2po}
\centering
\scriptsize
\setlength{\tabcolsep}{4pt}
\renewcommand{\arraystretch}{1.08}
\begin{tabular}{@{}p{0.68\linewidth}p{0.26\linewidth}@{}}
\toprule
\textbf{Hyperparameter} & \textbf{Value} \\
\midrule
Hidden size & 256 \\
MLP layers & 2 \\
Discount $\gamma$ & 0.96 \\
GAE $\lambda$ & 0.95 \\
PPO clip $\epsilon$ & 0.2 \\
Target KL & 0.016 \\
Update epochs & 10 \\
Minibatches & 2 \\
Actor LR & $5.0\times 10^{-4}$ \\
Critic LR & $5.0\times 10^{-3}$ \\
Entropy coef. & 0.0 \\
Max grad norm & 10.0 \\
\midrule
Cost budget $d$ & 25 \\
Dual init ($\lambda_0$) & 0.1 \\
Dual step & $5.0\times 10^{-4}$ \\
\midrule
Top-$k$ reads ($k$) & 3 \\
Memory embed dim & 64 \\
Hazard lookahead $H$ & 8 \\
Hazard cost thresh. $\delta$ & 0.1 \\
Write penalty coef. & 0.001 \\
Hazard loss coef. & 0.5 \\
Adaptive threshold & True \\
Init. threshold $\tau$ & 0.10 \\
Target write rate $\rho^\star$ & 0.05 \\
Threshold LR & 0.05 \\
Threshold bounds & [0.05, 0.95] \\
\bottomrule
\end{tabular}
\end{subtable}\hfill
\begin{subtable}[t]{0.48\textwidth}
\caption{\textsc{MAPPO-Lag}}
\label{tab:hparams-mappolag}
\centering
\scriptsize
\setlength{\tabcolsep}{4pt}
\renewcommand{\arraystretch}{1.08}
\begin{tabular}{@{}p{0.68\linewidth}p{0.26\linewidth}@{}}
\toprule
\textbf{Hyperparameter} & \textbf{Value} \\
\midrule
Hidden size & 256 \\
MLP layers & 2 \\
Discount $\gamma$ & 0.96 \\
GAE $\lambda$ & 0.95 \\
PPO clip $\epsilon$ & 0.2 \\
Target KL & 0.016 \\
Update epochs & 10 \\
Minibatches & 2 \\
Actor LR & $9.0\times 10^{-5}$ \\
Critic LR & $5.0\times 10^{-3}$ \\
Entropy coef. & 0.0 \\
Max grad norm & 10.0 \\
\midrule
Cost budget $d$ & 25 \\
Dual init ($\lambda_0$) & 0.78 \\
Dual step & $1.0\times 10^{-5}$ \\
\bottomrule
\end{tabular}
\end{subtable}
\end{table}
\vfill


\clearpage
\begin{table}[t]
\caption{\textbf{Key method hyperparameters (2/3).} Common run settings are in Table~\ref{tab:common-run-settings}.}
\label{tab:hparams-page2}
\centering
\begin{subtable}[t]{0.48\textwidth}
\caption{\textsc{MAPPO}}
\label{tab:hparams-mappo}
\centering
\scriptsize
\setlength{\tabcolsep}{4pt}
\renewcommand{\arraystretch}{1.08}
\begin{tabular}{@{}p{0.68\linewidth}p{0.26\linewidth}@{}}
\toprule
\textbf{Hyperparameter} & \textbf{Value} \\
\midrule
Hidden size & 256 \\
MLP layers & 2 \\
Discount $\gamma$ & 0.96 \\
GAE $\lambda$ & 0.95 \\
PPO clip $\epsilon$ & 0.2 \\
Target KL & 0.016 \\
Update epochs & 10 \\
Minibatches & 2 \\
Actor LR & $9.0\times 10^{-5}$ \\
Critic LR & $5.0\times 10^{-3}$ \\
Entropy coef. & 0.0 \\
Max grad norm & 10.0 \\
\bottomrule
\end{tabular}
\end{subtable}\hfill
\begin{subtable}[t]{0.48\textwidth}
\caption{\textsc{MACPO}}
\label{tab:hparams-macpo}
\centering
\scriptsize
\setlength{\tabcolsep}{4pt}
\renewcommand{\arraystretch}{1.08}
\begin{tabular}{@{}p{0.68\linewidth}p{0.26\linewidth}@{}}
\toprule
\textbf{Hyperparameter} & \textbf{Value} \\
\midrule
Hidden size & 256 \\
MLP layers & 2 \\
Discount $\gamma$ & 0.96 \\
GAE $\lambda$ & 0.95 \\
PPO clip $\epsilon$ & 0.2 \\
Target KL & 0.016 \\
Update epochs & 10 \\
Minibatches & 2 \\
Actor LR & $9.0\times 10^{-5}$ \\
Critic LR & $5.0\times 10^{-3}$ \\
Entropy coef. & 0.0 \\
Max grad norm & 10.0 \\
\midrule
Cost budget $d$ & 25 \\
Safety $\gamma$ & 0.09 \\
Step fraction & 0.5 \\
CG iters & 10 \\
$g$-step dir coef. & 0.1 \\
$b$-step dir coef. & 0.1 \\
Fraction coef. & 0.1 \\
EPS & $1.0\times 10^{-8}$ \\
\bottomrule
\end{tabular}
\end{subtable}
\end{table}


\clearpage
\begin{table}[t]
\caption{\textbf{Key method hyperparameters (3/3)} Common run settings listed here.}
\label{tab:hparams-page3}
\centering
\begin{subtable}[t]{0.48\textwidth}
\caption{\textsc{HAPPO}}
\label{tab:hparams-happo}
\centering
\scriptsize
\setlength{\tabcolsep}{4pt}
\renewcommand{\arraystretch}{1.08}
\begin{tabular}{@{}p{0.68\linewidth}p{0.26\linewidth}@{}}
\toprule
\textbf{Hyperparameter} & \textbf{Value} \\
\midrule
Hidden size & 256 \\
MLP layers & 2 \\
Discount $\gamma$ & 0.96 \\
GAE $\lambda$ & 0.95 \\
PPO clip $\epsilon$ & 0.2 \\
Target KL & 0.016 \\
Update epochs & 10 \\
Minibatches & 2 \\
Actor LR & $5.0\times 10^{-4}$ \\
Critic LR & $5.0\times 10^{-4}$ \\
Entropy coef. & 0.0 \\
Max grad norm & 10.0 \\
\bottomrule
\end{tabular}
\end{subtable}\hfill
\begin{subtable}[t]{0.48\textwidth}
\caption{\textbf{Common run settings (all methods)}}
\label{tab:common-run-settings}
\centering
\scriptsize
\setlength{\tabcolsep}{4pt}
\renewcommand{\arraystretch}{1.08}
\begin{tabular}{@{}p{0.68\linewidth}p{0.26\linewidth}@{}}
\toprule
\textbf{Setting} & \textbf{Value} \\
\midrule
Total environment steps & $3{,}000{,}000$ \\
Num parallel environments (\texttt{--num-envs}) & 16 \\
Num seeds (for uncertainty bands) & 3 \\
\bottomrule
\end{tabular}
\end{subtable}
\end{table}

\end{document}